\documentclass[conference]{IEEEtran}
\IEEEoverridecommandlockouts
\usepackage{tabularray}
\usepackage{cite}
\usepackage{amsmath,amssymb,amsfonts}
\usepackage{algorithmic}
\usepackage{graphicx}
\usepackage{textcomp}
\usepackage{xcolor}
\def\BibTeX{{\rm B\kern-.05em{\sc i\kern-.025em b}\kern-.08em
    T\kern-.1667em\lower.7ex\hbox{E}\kern-.125emX}}
\begin{document}

\onecolumn
\textbf{IEEE Copyright Notice}

© 2023 IEEE.  Personal use of this material is permitted.  Permission from IEEE must be obtained for all other uses, in any current or future media, including reprinting/republishing this material for advertising or promotional purposes, creating new collective works, for resale or redistribution to servers or lists, or reuse of any copyrighted component of this work in other works.
\twocolumn
\newpage

\title{Machine Learning on Dynamic Graphs: A Survey on Applications
}

\author{\IEEEauthorblockN{1\textsuperscript{st} Sanaz Hasanzadeh Fard}
\IEEEauthorblockA{\textit{dept. Computer Science and Engineering} \\
\textit{Michigan State University}\\
East Lansing, Michigan \\
hasanzad@msu.edu}
% \and
% \IEEEauthorblockN{2\textsuperscript{nd} Emily Dolson}
% \IEEEauthorblockA{\textit{dept. Computer Science and Engineering} \\
% \textit{Michigan State University}\\
% East Lansing, Michigan \\
% dolsonem@msu.edu}

}

\maketitle

\begin{abstract}
Dynamic graph learning has gained significant attention as it offers a powerful means to model intricate interactions among entities across various real-world and scientific domains. Notably, graphs serve as effective representations for diverse networks such as transportation, brain, social, and internet networks. Furthermore, the rapid advancements in machine learning have expanded the scope of dynamic graph applications beyond the aforementioned domains. In this paper, we present a review of lesser-explored applications of dynamic graph learning. This study revealed the potential of machine learning on dynamic graphs in addressing challenges across diverse domains, including those with limited levels of association with the field.
\end{abstract}


\section{Introduction}
In recent years, there has been a growing interest in dynamic graph learning \cite{b1, kazemi2020representation, huang2023temporal, fu2022federated, rossi2020temporal, chang2020continuous, yamada2020time, manessi2020dynamic} which provides a powerful framework for modeling complex interactions among entities in various domains. The term \textit{dynamic graph learning} or more generally graph learning has been referred to in various words in the literature; dynamic graph learning \cite{you2022roland, chang2020continuous}, dynamic machine learning on graphs \cite{b1}, machine learning on temporal networks \cite{liu2021inductive}, dynamic network embedding/ learning \cite{xue2022dynamic}, and dynamic graph representation learning \cite{goyal2020dyngraph2vec, kazemi2020representation}, all mention the same concept; also, the term \textit{network science} and \textit{complex network analysis} refer to study large-scale networks with different tools from graph theory to machine learning \cite{barabasi2013network, costa2007characterization}. Continues-time link prediction \cite{qu2020continuous}, temporal link prediction \cite{divakaran2020temporal, lei2019gcn, qin2022temporal, qiu2021temporal, yang2019advanced, ma2022joint}, and dynamic link prediction \cite{lei2019gcn} are the known words in the study of the applications of machine learning on dynamic graphs, but these applications are not limited to link prediction tasks. Graphs have proven to be effective representations for a wide range of networks, including health-related networks such as for Alzheimer analysis \cite{zhu2022interpretable} and drug discovery \cite{gaudelet2021utilizing}, epidemiology analysis as for Covid-19 \cite{la2020epidemiological}, transportation and traffic networks \cite{peng2021dynamic, peng2020spatial, li2023spatial, hu2022graph}, brain networks\cite{kim2021learning}, multi-agent environments \cite{hasanzadeh2019two}, and social networks \cite{xue2022dynamic, kazemi2020representation, knoke2019social}. The advancements in machine learning have expanded the applications of dynamic graph learning beyond these traditional domains. In order to gain a comprehensive understanding of machine learning on graphs, it is important to delve into the nuances of this field beyond a high-level overview. Machine learning algorithms are designed to solve problems by developing models through the use of training data, which can then be applied to test data \cite{janiesch2021machine}. However, these algorithms are typically designed to handle real-valued data and cannot be directly applied to graph problems due to their special structure. In order to address this issue, there are two main approaches to graph learning: embedding graphs into a latent real-valued space and applying machine learning models to the embeddings, or generating models that can be directly applied to graph-structured data (such as graph neural networks) \cite{chami2022machine}. In the case of time-evolving graphs, machine learning on dynamic graphs should also capture the evolution of networks over time.
 
In this paper, we aim to present a review of lesser-explored applications of dynamic graph learning. While much of the existing research has focused on well-known domains, such as transportation and social networks, there are areas with less explicit correlation with the field but they can highly benefit from dynamic graph analysis; machine vision applications such as object tracking, air quality analysis, and construction applications like clogging detection are examples. By exploring these unexplored applications, we hope to uncover new opportunities and challenges in the field of dynamic graph learning.

The rest of this paper is structured as follows: the second and third sections provide an overview of how dynamic graph learning approaches function, discussing data models and the general technical direction. The fourth section explores six specific applications in various domains that have employed dynamic graph learning to address problems related to multiple target tracking, air quality prediction, clogging detection, anomaly detection, bus load forecasting, and anti-money laundering. Finally, the conclusion summarizes the key findings of the paper.

\section{Data Model}
Dynamic graph learning approaches can be categorized in different ways. One criterion for classifying these approaches is the data model they employ. Dynamic graph learning data models fall into two groups: discrete models and continuous ones \cite{xue2022dynamic}. These two models represent the temporal evolution of networks over time. In the discrete-time model, there is a starting point of time and time intervals. In this model, there is a snapshot of the network at each point in time and each snapshot represents the state of the network, nodes, and connections between them. Each snapshot may be different from the previous and next snapshots. One example of a network that changes at discrete points in time is a social network where nodes (e.g. people) are added/ deleted (e.g. join/ leave) to/ from the network at each time; they also make connections or remove previously made ones. In the continuous model, the representation of the network is in the form of a continuous-time evolving network. In networks with continuous-time, changes occur in continuous time rather than discrete points. An example of such a situation is sensor networks. In these networks, sensor readings are constantly updated. 

When it comes to choosing a data model, each approach adopts a data model based on the underlying network and the type of changes in nodes and edges. It should be mentioned that some networks can be modeled in both ways. The advantage of the discrete model over the continuous model is that the implementation of the discrete model is simpler and also can be adopted from static network learning approaches by making some changes. Continuous time models on the other hand have the advantage of capturing complex dynamics and modeling temporal evolution more precisely.

\section{General Techniques}
Besides the classification of dynamic graph learning approaches based on data models, these approaches are classified based on the underlying technique they use \cite{kazemi2020representation}. Each method falls under one or a combination of the following techniques: 1- Temporal Graph Embeddings, 2- Recurrent Neural Networks (RNNs), 3- Graph Convolutional Networks (GCNs), 4- Dynamic Bayesian Networks (DBNs), 5- Matrix Factorization, and 6- Deep Reinforcement Learning.

The first class of dynamic graph learning approaches is temporal graph embedding. Graph embedding is a technique of representing graphs by mapping nodes and edges of the graph to a latent real-valued space and representing them as fixed-dimension vectors \cite{cai2018comprehensive, xu2021understanding, goyal2018graph, yan2006graph}. In temporal graph embedding the mapping process to the low-dimensional space involves preserving nodes' and edges' temporal relationship \cite{barros2021survey, xu2021understanding, goyal2020graph, wang2022survey}. Dynamic graph embedding approaches span into two sub-classes: auto-encoders \cite{mahdavi2020dynamic, xuan2022dynamic} and random walk-based \cite{jabri2020space, carletti2020random, xia2019random} approaches. As mentioned before, this series of approaches capture the temporal dynamics of the network by transferring them into a low-dimensional vector space. These real-valued vectors can be fed to common machine-learning models for further analysis. 

The second series of approaches are recurrent neural networks \cite{yu2019review, staudemeyer2019understanding, wang2022predrnn}. A recurrent neural network (RNN) is a type of artificial neural network that is designed to process sequential data, such as time-series data or natural language sentences. Unlike traditional feedforward neural networks that process input data in a single pass, RNNs have a feedback loop that allows them to maintain an internal state or memory of previous inputs. This enables RNNs to capture the temporal dependencies in sequential data and make predictions based on the context of previous inputs. In the context of temporal graph analysis, the dynamic network is the input to RNNs as a time series; RNNs capture temporal dependencies in the input network; the captured dependencies can be short-term or long-term based on the used neural network. RNNs work with sequential data and directly model the dynamics of the network. 

The third series of approaches are graph convolutional networks (GCN) \cite{zhang2019graph, chen2020simple, bo2021beyond, wu2019simplifying}. Analyzing graph-structured data is a challenging task due to their complex structure and ordinary neural networks cannot be applied to graph data; A graph convolution network is a type of neural network that is designed to operate on graph-structured data. Unlike traditional convolutional neural networks that operate on grid-structured data, GCNs use graph convolution operations to aggregate information from neighboring nodes in a graph. This enables GCNs to capture the structural relationships between nodes in a graph and make predictions based on the local and global structure of the graph. 

The fourth series of dynamic graph learning techniques are Dynamic Bayesian Networks (DBNs). DBNs \cite{scanagatta2019survey} are probabilistic graphical models that are used to represent and reason systems that change over time. They are a type of Bayesian network that incorporates temporal dependencies between variables, allowing them to model the evolution of a system over time. DBNs consist of a set of nodes representing variables and a set of directed edges representing the conditional dependencies between the variables at different time steps. Temporal link prediction as a task in the field of dynamic graph learning can be done by modeling the underlying probabilistic dependencies between nodes and edges in a network over time. The variables correspond to the nodes and edges in the network, and the DBN model captures the dependencies between them over time. The model can be learned from historical data using various algorithms such as Expectation-Maximization (EM) or Markov Chain Monte Carlo (MCMC) methods \cite{carlin1995bayesian}. Once the DBN model is learned, it can be used to make predictions about future links by computing the posterior probability distribution of the links given the observed data up to a certain time point. The predicted links can then be ranked based on their posterior probabilities, and the top-ranked links can be considered as the most likely to occur in the future. DBN technique provides a powerful framework for modeling the temporal dynamics of complex networks and making accurate predictions about future links based on probabilistic reasoning.

The next series of used techniques for dynamic graph learning is matrix factorization. Matrix factorization \cite{koren2009matrix, mehta2017review, baltrunas2011matrix, jamali2010matrix} is a mathematical technique used to decompose a matrix into two or more matrices that can be multiplied together to reconstruct the original matrix. The goal of matrix factorization is to find a low-dimensional representation of the original matrix that captures its essential features and allows for efficient computation. Matrix factorization can be done using various algorithms, such as singular value decomposition (SVD) \cite{baker2005singular}, non-negative matrix factorization (NMF) \cite{lee2000algorithms}, and principal component analysis (PCA) \cite{bro2014principal}. In the case of dynamic graphs, Matrix Factorization techniques decompose the adjacency matrix of a dynamic network into low-rank matrices that capture the underlying structure of the network. In this approach, the adjacency matrix of the graph is factorized into two low-rank matrices representing the latent features of the nodes and the latent features of the time periods. The latent features of the nodes capture their underlying properties, while the latent features of the time periods capture the temporal dynamics of the graph. The factorization is performed using optimization techniques such as stochastic gradient descent or alternating least squares. Once the factorization is complete, it can be used to predict future links by estimating the missing entries in the adjacency matrix. For example, if two nodes have similar latent features at a particular time period, it is likely that they will have a link in the future. MF-based approaches for temporal link prediction include methods such as Temporal Regularized Matrix Factorization (TRMF) \cite{yu2016temporal} and Dynamic Matrix Factorization (DMF) \cite{ma2018graph}.

The last series of dynamic graph learning techniques are Deep Reinforcement Learning techniques. Deep reinforcement learning \cite{li2017deep, arulkumaran2017deep, franccois2018introduction, henderson2018deep} is a subfield of machine learning that combines deep learning with reinforcement learning. It involves training an artificial intelligence agent to make decisions based on feedback from its environment. The agent learns through trial and error, receiving rewards for making good decisions and punishments for making bad ones. Deep reinforcement learning techniques involve using neural networks, which are trained to predict the best action for a given state. These networks are trained using a combination of supervised and unsupervised learning, and the reinforcement signal is used to update the weights of the neural network. In the context of dynamic graph learning, deep reinforcement learning techniques learn policies for controlling the evolution of a dynamic network. These policies can be used to optimize network performance or achieve specific objectives. For example, Deep Reinforcement Learning (DRL) technique can perform temporal link prediction by learning a policy that maximizes the expected cumulative reward over a sequence of actions; the actions correspond to predicting whether a link will exist or not at a future time step given the current state of the network. The state of the network can be represented by various features such as node attributes, edge attributes, and graph structure. The DRL agent interacts with the network by selecting actions based on its current state and receives a reward signal based on the accuracy of its predictions. The reward signal can be designed to encourage the agent to make accurate predictions while also penalizing it for making incorrect predictions. The agent then updates its policy based on the observed rewards and continues to interact with the network until convergence. One popular DRL algorithm for temporal link prediction is Deep Q-Network (DQN), which uses a deep neural network to approximate the Q-value function that represents the expected cumulative reward for each action in each state. Another popular algorithm is Actor-Critic, which uses two neural networks to approximate the policy and value functions, respectively. Overall, DRL techniques can effectively capture the temporal dynamics of complex networks and make accurate predictions about future links.

These different categories of approaches provide a range of options for modeling and predicting the behavior of dynamic networks using machine learning techniques.

\section{Diverse Applications of Dynamic Graph Learning Across Different Domains}
In this section, we discuss less-represented applications that adopted dynamic graph learning. The goal of this section is beyond the reviewed applications; we want to broaden the perspective of the capabilities of dynamic graph learning in providing solutions for various tasks. Table 1 demonstrates 6 tasks that dynamic graph learning has provided solutions for them as well as the underlying technique they used. These 6 tasks are chosen based on recent advancements which have applied dynamic graph learning as their original method of solving a problem. This study has special attention to works that not only applied dynamic graph learning to their problem-solving process but also proposed a new approach for dynamic graph learning; so, this work is also a review of recently proposed approaches for dynamic graph learning. These tasks cover a wide range of applications in both everyday life and science; we tried to select less abstract applications and also cover applications with less direct connection to the field, with the aim of representing dynamic graph learning's potential in providing solutions for a wider range of tasks. The rest of the section has studied these approaches in more detail.

\subsection{TransMOT: Spatial-Temporal Graph Transformer for Multiple Object Tracking}
Image processing and machine vision \cite{sonka2014image} provide solutions for a broad range of tasks including Multiple Object Tracking. The task of multiple objects (i.e. target) tracking \cite{luo2021multiple} is monitoring objects as they move. While this task can be seen as a pure machine vision task, TransMOT \cite{chu2023transmot} has proposed a solution for that in terms of dynamic graphs. TransMOT has focused on modeling the objects' spatial-temporal interactions using graph transformers \cite{min2022transformer} to track multiple objects in videos. 

TransMOT defines objects as a temporal series of sparse weighted graphs that are constructed using their spatial relationships within each frame. This way, TransMOT is able to detect and track a large number of objects more effectively. TransMOT represents objects in their environment as graph-structured data and encodes their features and relationships by Spatial-Temporal Graph Transformer Encoder \cite{meinhardt2022trackformer, sun2020transtrack}. The spatial graph transformer decoder models spatial and appearance correlations of detection candidates. TransMOT learns spatial-temporal association and generates the assignment matrix for multiple object tracking (i.e. MOT). 

\begin{table}
\centering
\caption{Studied models, their applications, and underlying techniques}
\begin{tblr}{
  width = \linewidth,
  colspec = {Q[204]Q[485]Q[244]},
  cells = {c},
  hlines,
  vlines,
}
\textbf{Model}                    & \textbf{Underlying Technique}                                                                                            & \textbf{Task}                                       \\
TransMOT                          & Autoencoder, GCN, Graph Multi-head Attention                                                                             & Multiple Object Tracking                            \\
BGGRU                             & Graph-SAGE, LSTM, Bayesian Graph GRU                                                                                     & Air Quality Prediction                              \\
{Explainable\\Spatio-temporal GCN} & GCN, Geological Eigenvector                                                                                              & Clogging Detection                                  \\
MST-GAT                           & Graph Attention Network, Temporal Conv.                                                                                  & Anomaly Detection                                   \\
ST-GCN                            & GCN, GRU, Spatial Conv. Layer                                                                                            & Bus Load Prediction                                 \\
Temporal-GCN                      & {LSTM, Topology Adaptive GCN (TAGCN)\\Bayesian Approximations (Monte-Carlo dropout and\\Monte-Carlo adversarial attack)} & {Anti-money Laundering\\(Bitcoin, Blockchain Net.)} 
\end{tblr}
\end{table}

TransMOT consists of a spatial graph transformer encoder layer, a temporal transformer encoder layer, and a spatial graph transformer decoder layer; using this structure, TransMOT models the interaction of objects by arranging the trajectories of the tracked targets and detection candidates as a set of sparse weighted graphs. TransMOT utilizes its ability to model a vast number of potential candidates and identify spatial-temporal clues to associate relevant candidates from a large pool of detection predictions. The majority of these predictions can be eliminated through detector post-processing.
% "Relying on the discriminative spatial-temporal clues and capability of modeling a large number of candidates, TransMOT can associate candidates from a large number of loosely filtered detection predictions, most of which can be discarded by the post-processing of a detector" \cite{chu2023transmot}.

TransMOT has been evaluated on multiple benchmark datasets including
MOT16 \cite{milan2016mot16} and MOT20 \cite{dendorfer2020mot20}, and achieved state-of-the-art performance on all the datasets. The importance of TranMOT is not limited to multiple object-tracking tasks; TransMOT demonstrates the potential of dynamic graph learning in addressing machine vision and image-processing tasks and can be inspiring for other applications in this field.

\subsection{Deep Spatio-Temporal Graph Network with Self-Optimization
for Air Quality Prediction}
Air pollution is one of the severe issues in today's life. Machine learning techniques are helpful in predicting the pattern of pollutants \cite{liang2020machine, zhang2022deep, li2016deep, zhu2018machine}; while time series analysis is a common approach to predicting the evolution patterns of air pollutants, these approaches ignore the spatial transmission effect of adjacent areas. Taking spatial transmissions into account can highly improve the accuracy of results. \cite{jin2023deep} has proposed a time series prediction network with the self-optimization ability of a spatio-temporal graph neural network (BGGRU) to mine the changing pattern of the time series and the spatial propagation effect. The proposed approach uses Graph-SAGE \cite{hamilton2017inductive} for graph sampling and aggregating to extract spatial information. In order to extract temporal dependencies, the proposed approach adopts a Bayesian Graph Gated Recurrent Unit (BGraphGRU). BGraphGRU predicts dynamics by applying a graph network to GRU. To improve the accuracy, the setting of the model hyperparameters is optimized using Bayesian Optimization. The high accuracy of the proposed approach by \cite{jin2023deep} has been verified by PM2.5 data (i.e. particles that are 2.5 microns or less in diameter \cite{california2020inhalable}).

\subsection{Learning from Explainable Data-Driven Tunneling Graphs: A Spatio-Temporal Graph Convolutional Network for Clogging Detection}
Tunneling shields have been widely used in the construction of underground structures. One of the common challenges in this field is clogging caused by mud cake. Mud cake can stick on the cutters and cement the rotary junction of the main bearing of the cutter head. To overcome challenges caused by these, clogging detection is a crucial part of the tunneling process. Machine learning models offer solutions for a wide range of applications and clogging detection \cite{bai2021pipejacking} based on real-time shield monitoring data \cite{qin2023novel, zhai2022clogging} is one of them. Deep learning models that use historical data offer promising solutions for clogging detection problems; the explainability issue of deep learning models still keeps research in this field ongoing. 

Spatio-temporal tunneling graphs and graph convolutional network models have been applied to the clogging detection problem as a response to the explainability issue. The graph modeling process in this regard as proposed in \cite{gao2023learning} is as follows: time-series monitoring data of tunneling shields are transferred as graphs. The definition of each node and edge in this graph modeling is a type of monitoring parameter and the partial correlation between monitoring parameters, respectively. This complex network model falls into the small-world property. 
% REF has studied the efficiency problem of machine learning models that do clogging detection from the explainability view. The proposed model is a spatiotemporal graph convolutional network for clogging detection. 
By calculating the partial correlation among thirty-two types of monitoring parameters of one ring (i.e. a slice of a thickened cylinder between two parallel planes), these parameters (as nodes) have been transformed into a tunneling graph. Graph embedding is learned by a graph convolution neural network based on hierarchical pooling and the attention mechanism; this model is used in order to determine the risk level of clogging for each ring. The proposed approach in \cite{gao2023learning} has been evaluated in multiple scenarios and experiments have proven the efficiency of the approach (e.g. the accuracy of high- and low-risk reached 90.75\%).

\subsection{MST-GAT: A Multimodal Spatial-Temporal Graph Attention Network for Time Series Anomaly Detection}
Multimodal Time Series (MTS) anomaly detection \cite{guo2018multidimensional}, as a special type of anomaly detection \cite{chandola2009anomaly, chalapathy2019deep, pang2021deep}, is monitoring diverse modalities of sensors in industrial devices and information technology systems; in this detection system, the data stream from each sensor is seen as a univariate time series. While monitoring each modality independently is not helpful, multimodal time series data aim to detect anomalies in the complex environment. Multimodal time series anomaly detection also benefits from early detection before all the modalities get damaged. Multimodal time series anomaly detection is a challenging task due to the complex spatial dependence (e.g., topological structure and modal correlation) and temporal dependence (e.g., period and trend) of multimodal time series. This correlation is not limited to time series from the same modality (referred to as intra-modal correlations), the correlation is also between time series from different modalities (referred to as inter-modal correlations).

Previous approaches \cite{zhang2003time, shipmon2017time, hill2007real, kromanis2013support} in the field of multimodal time series anomaly detection, capture temporal dynamics; the newer methods also try to consider the spatial dependence between different time series. The proposed approach in \cite{ding2023mst} has covered previous challenges and also captured the multimodal correlation among multimodal time series. In order to provide interpretable results, the proposed approach locates a univariate time series for each anomaly that is the most likely cause of that anomaly. \cite{ding2023mst} has proposed a multimodal spatial-temporal graph attention network, named MST-GAT. This network explicitly captures modal dependencies between multimodal time series using the prevalent graph attention networks (GATs). This multimodal graph attention network (M-GAT) includes a multi-head attention module and two relational attention modules, i.e., intra- and inter-modal attention, to capture the spatial dependencies between multimodal time series. 

This model benefits from a good feature representation by explicitly modeling different relationships in multimodal time series. Then a temporal convolution network captures temporal dependencies in each time series with a standard convolution operation on time slices. Moreover, this model benefits from the joint optimization of a reconstruction module and a prediction module to integrate their advantages. The role of the reconstruction module is to reconstruct the input data and the prediction module aims at predicting the feature of the next timestamp. \cite{ding2023mst} uses the reconstruction probability and the prediction error to explain the detected anomalies. MST-GAT has been evaluated on benchmark datasets including MSL \cite{hundman2018detecting}, SMAP \cite{hundman2018detecting}, SWaT \cite{goh2017dataset}, and WADI \cite{ahmed2017wadi} and has outperformed the state-of-the-art baselines. MST-GAT also strengthens the interpretability of detected anomalies by locating the most anomalous univariate time series.

\subsection{Gated Spatial-Temporal Graph Neural Network Based Short-Term Load Forecasting for Wide-Area Multiple Buses}

Bus load forecasting is a crucial task in people's lives. One example is during extreme weather conditions, such as heat waves or cold snaps. During these times, there is a significant increase in demand for electricity as people turn to air conditioning or heating systems to stay comfortable. Without accurate load forecasting, grid operators may struggle to meet the sudden increase in demand, leading to power outages and blackouts. Bus load forecasting in power systems helps bring safety to the grid dispatch operations and more accurate online analysis decisions. Bus load forecasting is a challenging task for multiple reasons: 1- the bus load is smaller than the system load, 2- the bus load is volatile and random because of the influence of meteorological factors and the behavior of users in the power supply area, and 3- changes in trends are not obvious. On the other side, due to factors such as differences in the power consumption behaviors of each bus supply object, bus loads can be different from each other resulting in difficulty in forecasting bus loads using a unified forecasting model. 

Machine learning-based approaches \cite{wang2020short} for short-term load forecasting are promising due to their capabilities in capturing complex nonlinear data features. GCNs provide a new technical route for diverse representation and feature mining of data within power systems. A combination of graph theory, graph convolutional neural networks, and rough sets have been used in \cite{khodayar2018spatio} to accurately forecast wind power. Previously proposed techniques do not consider the influence of meteorological and date factors on load fluctuations. The spatial convolution layer (SCL) of GCN directly processes graph structure data by extracting the spatial coupling relationships embedded in the bus load data which results in a reduction in the number of outlier points that considerably affect scheduling in bus load forecasting. 

The remaining challenge is that the spatial convolution layer of GCN mines the spatial domain features among bus loads, but the temporal domain features among the similar-weighted spatial–temporal graph still needs to be mined. \cite{huang2023gated} proposes a deep model structure to construct a short-term load forecasting method for wide-area multiple buses based on ST-GCN. This model constructs a multi-node feature set and the node feature types in the similar-weighted spatial–temporal graph are expanded to improve the accuracy of the worst evaluation metrics. In this modified model, the RapidMIC value determines the node connection relationship, and the RapidMIC value represents the edge feature in the similar-weighted spatial–temporal graph.

% The difference between the proposed model by the \cite{ref_article11} and the model of [22] is that \cite{ref_article11} constructs similar-weighted spatial-temporal graphs with weights. 
The constructed similar-weighted spatial-temporal graph in this model reflects the spatial coupling and correlation degree between the buses. This model and its results can be interpreted as follows: 1- edge features: RapidMIC value between bus loads, 2- the combination of the multiple node feature set constructs a similar-weighted spatial-temporal graph, 3- the spatial convolution layer is constructed to extract the high-dimensional features of neighboring nodes in the similar-weighted spatial-temporal graph and generate spatial aggregation features to achieve full-domain node feature enhancement, and 4- spatial aggregation features at different moments are constructed into temporal series and input into gated recurrent unit layer (GRUL) to mine its temporal domain features. Based on this, the wide-area multiple buses short-term load forecasting is achieved. This way, \cite{huang2023gated} addresses a unified model for accurate short-term load forecasting of multiple bus loads over a wide area. Compared to other methods, the proposed approach by \cite{huang2023gated} has improved evaluation metrics by 1.82\% to 5.94\% and it has better robustness in the scene with abnormal load data. 

\subsection{Graph-Based LSTM for Anti-money Laundering: Experimenting Temporal Graph Convolutional Network with Bitcoin Data}
One of the main criminal activities in the domain of Blockchain and Bitcoin is related to money laundering; reports revealed 4.5\$ billion in 2019 of Bitcoin crime related to illicit services. This fact has attracted the attention of many researchers to develop intelligent methods that exploit the transparency of blockchain records. These methods result in Anti-Money Laundering (AML) to improve safe cryptocurrency transactions. In this regard, researchers have benefited from a dataset provided by a cryptocurrency intelligence company focusing on safeguarding cryptocurrency systems (Elliptic Company). The mentioned dataset, called Elliptic data, is a graph network of Bitcoin transactions that acquires a graph of Bitcoin transactions that spans handcrafted local features (associated with transactions themselves) and aggregated features (associated with neighboring transactions) with partially labeled nodes. The labeled nodes also denote licitly-transacted payments (e.g. miners) and illicit transactions (e.g. theft, scams). 

This dataset has been used to verify techniques in the field of classical supervised learning methods \cite{weber2019anti, alarab2020comparative, alarab2020competence, pareja2020evolvegcn, poursafaei2021sigtran, alarab2021illustrative}. The drawback of these approaches is that they ignore the temporal information of this dataset. In order to fill the gap in the previously proposed approaches, \cite{alarab2023graph} has proposed a model by considering both structural and temporal information of Elliptic data into account in order to predict illicit transactions that belong to illegal services in the Bitcoin blockchain network. They have also invested effort in performing active learning on Bitcoin blockchain data that mimics a real-life situation in order to deal with massively growing Bitcoin blockchain technology and its challenging time-consuming process of labeling that. \cite{alarab2023graph} classification model uses LSTM and GCN models. The active learning framework should contain an acquisition function that relies on the model’s uncertainty to query the most informative data. The model’s uncertainty estimates are obtained using two comparable methods based on Bayesian approximations which are named Monte-Carlo dropout \cite{gal2016dropout} and Monte-Carlo adversarial attack \cite{alarab2022adversarial}. The proposed model has outperformed the previous studies with an accuracy of 97.77\% under the same experimental settings.

\section{Conclusion}
In this paper, we leveraged the importance of extending dynamic graph learning research to unrepresented tasks in this field which in-depth have the potential to be effectively addressed by graph learning. While the reviewed applications are not in the primary circle of problems that temporal graph learning tries to solve, we saw by effective definition they may be highly correlated with the graph tasks, and graph-related techniques can provide practical solutions for them. Through our research, we have discovered that dynamic graph learning holds great promise in tackling crucial challenges that impact people's daily lives; it has proven to be a valuable tool in predicting air quality and aiding construction efforts, as well as in analyzing global economic issues like anti-money laundering in the Blockchain network. Additionally, dynamic graph learning has demonstrated its potential in enhancing security and surveillance measures through tasks such as anomaly detection and object tracking. 
Research on dynamic graph learning is advancing to encompass real-time analysis of dynamic graphs and multi-modal dynamic graph analysis. The future direction of our research aims to unleash the potential of dynamic graph learning with real-time and multi-modal analysis in various domains, spanning from complex abstract concepts to everyday practical applications. This direction of research seeks to broaden the scope of dynamic graph analysis and expand its utility in diverse fields.

\section*{Acknowledgment}
We would like to thank Dr. Mohammad Ghassemi for their insightful comments that helped improve this manuscript.

\end{document}